\documentclass[11pt]{article}

\usepackage[final]{acl}

\usepackage{times}
\usepackage{latexsym}

\usepackage[T1]{fontenc}

\usepackage[utf8]{inputenc}

\usepackage{microtype}

\usepackage{inconsolata}

\usepackage{graphicx}

\usepackage{amsmath} 
\usepackage{amssymb}
\usepackage{booktabs} 
\usepackage{multirow} 
\usepackage{algorithm}
\usepackage{algorithmic}

\usepackage{xcolor}
\usepackage{listings}
\definecolor{codebg}{RGB}{248,248,248}

\title{AIA$^{2}$: Attribute-Agnostic Imbalance Augmentation for Subgroup Robustness}

\author{Hanshu Rao\\
  University of Memphis \\
  \texttt{hrao@memphis.edu} \\\And
  Guangzeng Han \\
  University of Memphis \\
  \texttt{ghan@memphis.edu} \\\And
  Xiaolei Huang \\
  University of Missouri \\
  \texttt{xiaolei.huang@missouri.edu} \\
  }

\begin{document}
\maketitle
\begin{abstract}
Attributes describing data content and context can induce diverse imbalance patterns that go beyond label imbalance alone.
However, existing studies primarily address label imbalance while overlooking data attributes, such as topics and demographics, which can induce meaningful subgroup structure while causing model degradation on underrepresented subgroups.
We propose Attribute-Agnostic Imbalance Augmentation (AIA$^{2}$)\footnote{Code is available at \url{https://github.com/trust-nlp/AIA2-Subgroup-Robustness}}, a framework for improving model robustness under varying subgroup imbalances without explicit subgroup annotations.
AIA$^{2}$ automatically discovers varying imbalances via latent semantic distributions, obtains slices with both learning difficulty and subgroup imbalance deficits, and deploys a large language model (LLM) for subgroup-aware imbalance augmentation.
We have evaluated AIA$^{2}$ on 5 popular corpora with rich domains and their attribute values, covering social issues and diverse topics.
Results show improved performance on the lowest-performing subgroups and consistent gains over competitive baselines.
Ablation studies confirm complementary contributions from each component, and additional analyses show that AIA$^{2}$ provides a practical and consistent way to improve worst-group robustness under data subgroup imbalance.
\end{abstract}

\section{Introduction}

\begin{figure}[htp]
\centering
\includegraphics[width=\columnwidth]{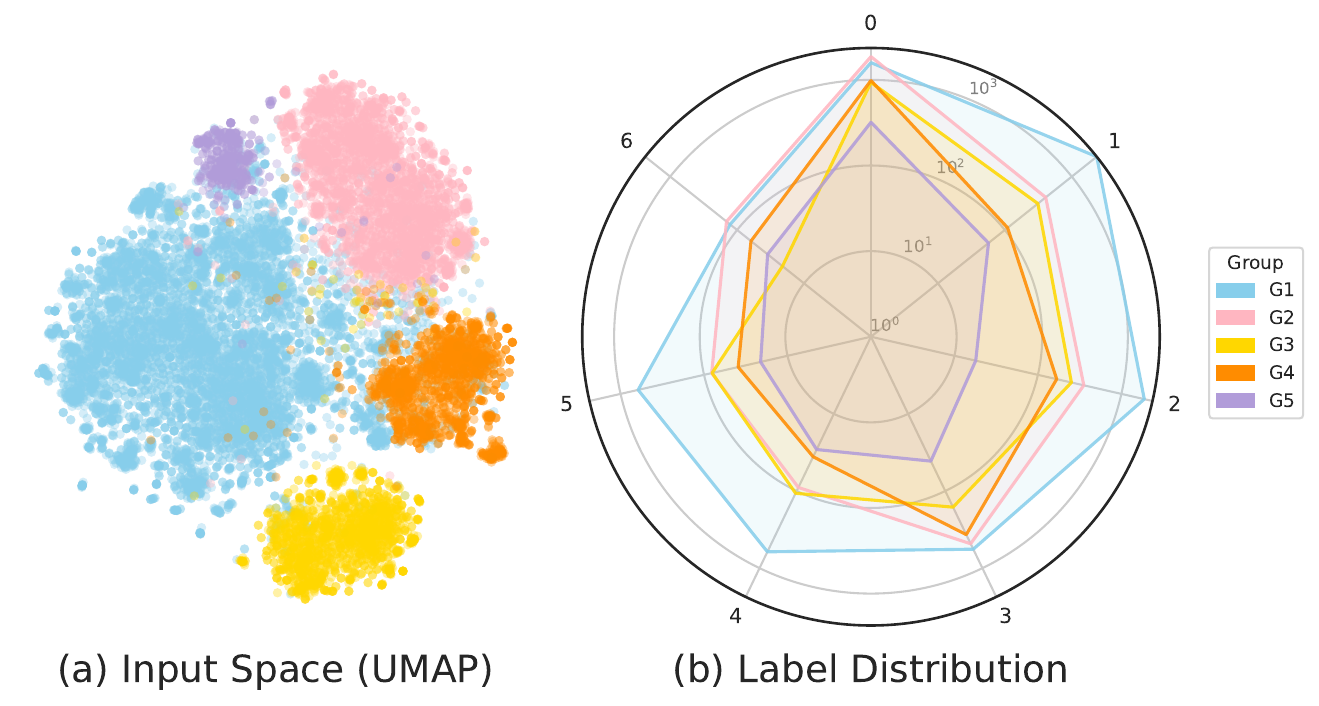}
\caption{Subgroup-level distribution shifts in HumAID across different event sources (colors).}
\label{fig:data_viz}
\end{figure}

Data attributes explicitly characterize data with heterogeneous attribute values and categorize data into subgroups with distinct statistical properties, such as topic, demographic factors, and various sources~\cite{WILDS_2021_Koh,Chen2023HiBug,li_joint_2025,liu-etal-2025-examining}.
Those data attributes can induce diverse patterns of subgroup imbalance that are not reflected by general data and label imbalance alone.
However, varying imbalance patterns by attribute values are usually overlooked in existing studies~\cite{rudner2024mind},
which can easily lead to unstable performance on underrepresented or minority subgroups and raise concerns on model robustness and reliability.
For example, a model optimized for overall label distributions may still exhibit significant performance variations or disparities across groups of data attributes with varying distributions.

\begin{figure*}[t]
\centering
\includegraphics[width=0.915\textwidth]{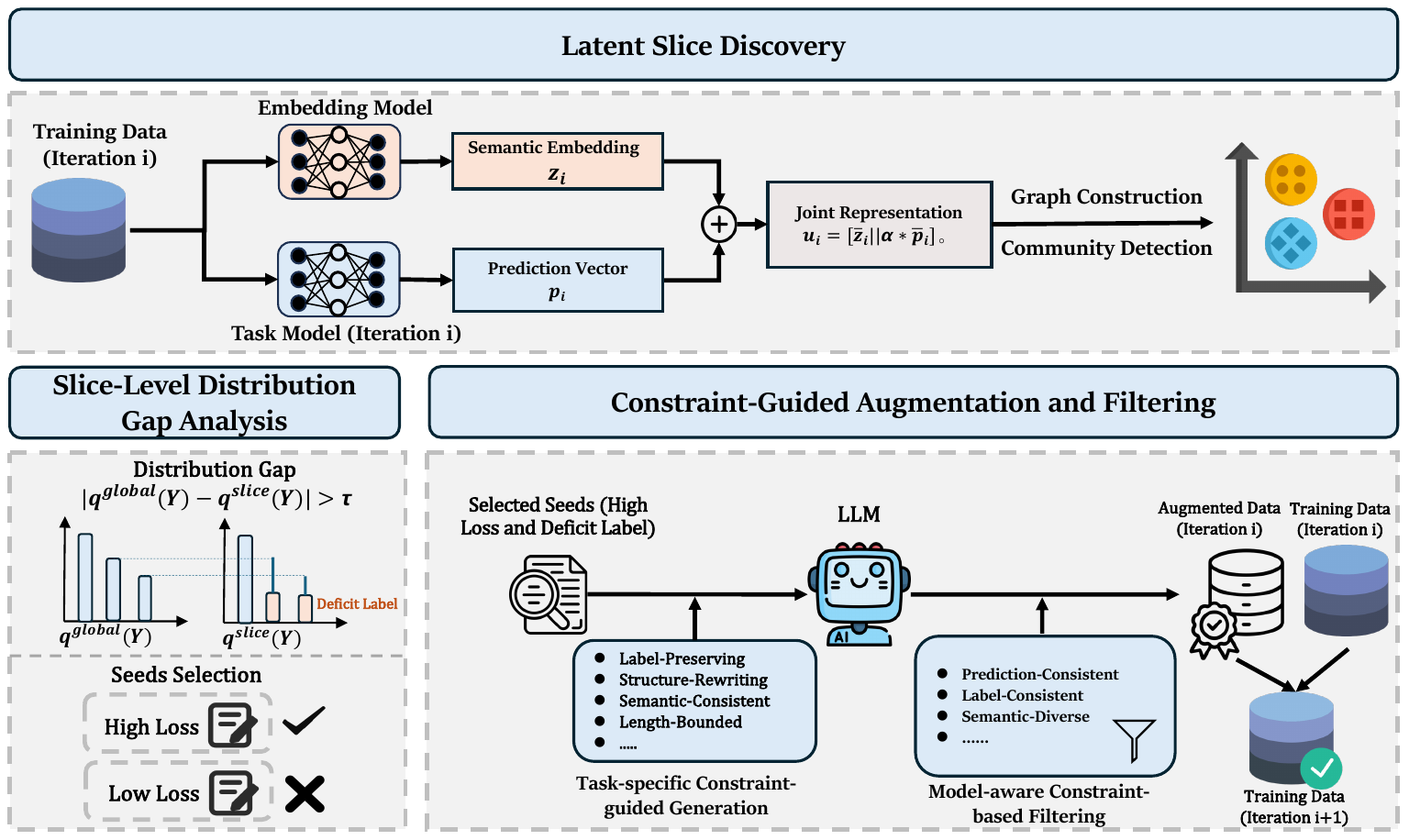}
\caption{
\textbf{Overview of the AIA$^{2}$ framework.}
(1) \textbf{Latent Slice Discovery} partitions examples into coherent slices using joint semantic-predictive representations; 
(2) \textbf{Slice-Level Distribution Gap Analysis} identifies deficit labels and selects high-loss seeds; 
(3) \textbf{Constraint-Guided Augmentation and Filtering} generates and filters candidates for consistency and diversity. 
Augmented data is merged into the training set for the next iteration.
}
\label{fig:method}
\end{figure*}

Figure~\ref{fig:data_viz} visualizes social media subgroup distributions by event source and label distributions: label imbalance widely exists across the subgroups, while imbalance patterns vary significantly across event sources.
While a growing body of work focuses on data balance through re-weighting~\cite{lin2017focal, rao2026model}, sampling~\cite{chawla2002smote}, data augmentation~\cite{zhang2023deep}, and optimization~\cite{ren2020balanced}, these studies usually consider only global-level label skew, leaving subgroup-level data skew underexplored.
Additionally, existing approaches~\cite{henning2023survey} implicitly or explicitly assume that attribute values and annotations are accessible; however, subgroup-defining attributes are often unknown, partially observed, sensitive, or unavailable.
For example, demographic attributes may be restricted due to privacy or regulatory constraints.
These gaps limit existing imbalance methods particularly when subgroup structure is implicit rather than explicit.

In this work, we introduce Attribute-Agnostic Imbalance Augmentation (AIA$^{2}$), a general framework for enhancing model robustness to subgroup imbalance induced by unknown or latent distributional variations.
Instead of predefined attribute values or explicit subgroup labels, AIA$^{2}$ automatically discovers latent slices with a joint semantic and learning space, identifies regions of distributional imbalance and misalignment, and constructs latent subgroup graphs to guide data and model augmentation.
AIA$^{2}$ employs distributional-gap-aware and constraint-guided data augmentation by large language models (LLMs) to dynamically select and fix subgroup imbalance deficits.
Because these latent slices evolve with the task model, AIA$^{2}$ can track localized failure regions without requiring predefined group assignments.
In sum, our contributions are threefold: 
1) we propose an attribute-agnostic imbalance-learning framework that improves worst-subgroup robustness without using subgroup metadata during training or model selection;
2) we present a subgroup-aware distributional gap analysis and imbalance-deficit-driven selection mechanism that targets minority patterns and labels within latent regions; 
and 3) we propose a constraint-guided filtering strategy and augmentation to ensure data diversity and label fidelity.

\section{Method}

Prior work on robustness under distribution shift and imbalance typically improves worst-case performance by reweighting or resampling training data across predefined groups, or by optimizing objectives that emphasize underperforming subpopulations~\cite{sagawa2019distributionally,Just2021Liu}. 
While these approaches effectively address global class imbalance or known group disparities, they implicitly assume that the relevant failure modes align with available metadata, either explicit group annotations or observable class boundaries. 
In practice, however, systematic errors often concentrate in latent regions where the local label distribution deviates sharply from the global one, and these regions evolve dynamically as the model learns, cutting across any predefined groupings.

This raises a fundamental question: \textit{when the model's worst failures occur in such unobserved, locally-misaligned regions, how can we identify where the distribution is ``broken'' and intervene without access to the right grouping?} 
To address this challenge, we propose an iterative slice-aware augmentation framework that 
i) discovers latent slices by clustering examples in a joint semantic-predictive space, 
ii) quantifies local distribution gaps to identify which labels are deficient in each slice, 
and iii) selectively generates targeted augmentations from the hardest examples while ensuring label consistency and semantic diversity. 
By iteratively repairing these local distribution defects, our method reduces concentrated slice-level errors without requiring known subgroups.
Figure~\ref{fig:method} shows the overview of our method.

\subsection{Latent Slice Discovery}

To identify where errors concentrate, we partition the data into slices based on two complementary signals: 
the input distribution $P(X)$ captured by semantic embeddings, and the predictive distribution $P(Y|X)$ captured by model outputs. 
Clustering on $P(X)$ alone groups semantically similar examples but ignores model behavior; 
clustering on $P(Y|X)$ alone captures prediction patterns but conflates unrelated inputs. 
By combining both, we discover regions where similar content elicits similar (often erroneous) predictions, the failure modes that single-distribution approaches miss.

We construct a joint representation space that captures both semantic content and predictive behavior. 
For each training example $i$, we combine its semantic embedding $\mathbf{z}_i \in \mathbb{R}^{d_z}$ (obtained from a pre-trained BGE encoder~\cite{bge_embedding}) with its prediction vector $\mathbf{p}_i$. 
For classification tasks, $\mathbf{p}_i \in \mathbb{R}^{C}$ corresponds to softmax outputs for single-label or sigmoid outputs for multi-label settings. 
For sequence labeling tasks such as named entity recognition, where predictions occur at the token level, we instead construct a boundary behavior vector $\mathbf{p}_i \in \mathbb{R}^{3C_E}$ for $C_E$ entity types.
This vector encodes, for each type, the average prediction loss on gold entity tokens, the entropy at boundary regions including adjacent positions, and the negative margin between correct and competing label probabilities—collectively capturing how reliably the model identifies entity boundaries rather than aggregating raw token-level outputs.
The unified representation is defined as:
\begin{equation*}
\mathbf{u}_i = [\bar{\mathbf{z}}_i \,||\, \alpha \cdot \bar{\mathbf{p}}_i]
\end{equation*}
, where $\bar{\mathbf{z}}_i$ and $\bar{\mathbf{p}}_i$ are L2-normalized vectors, $||$ denotes concatenation, and $\alpha$ controls the relative importance of predictive information. 
This design ensures that slices capture coherent regions in the joint space of 
$P(X)$ (via $\mathbf{z}_i$) and $P(Y|X)$ (via $\mathbf{p}_i$), 
rather than relying on either signal alone.

To discover coherent slices from this joint space, we construct a k-nearest neighbor graph where edges connect similar examples, then apply the Leiden community detection algorithm~\cite{traag2019leiden} with resolution parameter $\gamma$ to control granularity. 
Unlike hierarchical clustering, Leiden optimization naturally handles varying cluster sizes and discovers communities at multiple scales, making it well-suited for finding both broad patterns and concentrated error pockets. 
The resulting slices provide a data-driven partitioning evolving with the model's behavior across training iterations.

\subsection{Slice-Level Distribution Gap Analysis}

Having discovered slices that are coherent in $P(X) \times P(Y|X)$ space, we now ask: \emph{which labels should we augment within each slice?} The answer lies in the gap between a slice's local label distribution $P^{(s)}(Y)$ and the empirical global distribution $P^{\text{global}}(Y)$.
For each slice $s$, we compute the local label distribution $\mathbf{q}^{(s)}$ and compare it with the global distribution $\mathbf{q}^{(\text{global})}$.
For multi-class classification, these are simply the empirical class frequencies; for multi-label tasks, they represent per-label positive rates. 
For sequence labeling tasks such as NER, we define distributions at the span level rather than the token level—$\mathbf{q}^{(s)}_c$ represents the proportion of entity spans of type $c$ within slice $s$, computed by counting contiguous B/I segments for each entity type. 
This span-based formulation better reflects the semantic composition of entities and avoids bias from varying entity lengths. 
We quantify the distribution gap using L1 distance:
\begin{equation*}
\text{gap}_s = \|\mathbf{q}^{(\text{global})} - \mathbf{q}^{(s)}\|_1
\end{equation*}
When $\text{gap}_s > \tau$, we select slice $s$ for augmentation.
More importantly, we compute the \emph{deficit vector} $\mathbf{d}^{(s)} = \max(\mathbf{0}, \mathbf{q}^{(\text{global})} - \mathbf{q}^{(s)})$, which indicates which labels are under-represented relative to the global distribution and by how much.

To select seed examples for augmentation, we prioritize candidates based on both label deficits and individual difficulty.
For each deficit label $c$ with $d^{(s)}_c > 0$, we identify candidate examples in slice $s$ with label $c$, excluding any previously augmented samples.
We then rank candidates by their cross-entropy loss $\ell_i$, selecting the hardest examples as seeds. 
The number of seeds per label is proportional to its deficit mass, with the total augmentation budget for slice $s$ determined by:
\begin{equation*}
n_s = \min\left(n_{\max}, \, \lceil |S_s|^{\beta} \cdot \sum_c d^{(s)}_c \cdot \rho \rceil\right)
\end{equation*}
, where $S_s$ is a set of examples in the slice $s$, $\beta \in (0,1)$ provides sublinear scaling to prevent large slices from dominating, $\rho$ controls the overall augmentation rate, and $n_{\max}$ caps the per-slice budget.

\subsection{Constraint-Guided Augmentation and Filtering}

Given the deficit labels identified through $P(Y)$ gap analysis within each $P(X) \times P(Y|X)$-defined slice, we now generate augmentations that maintain label fidelity while introducing sufficient diversity to improve local coverage. 
We leverage a large language model (LLM) with carefully designed constraints to generate meaningful candidates.

For classification tasks, we design prompts that enforce structural transformation and entity substitution while preserving label-determining semantics. 
Each prompt requires 1) sentence-level restructuring beyond simple word swaps and 2) type-preserving entity replacement (e.g., person→person, location→location).
Sequence labeling tasks like NER present a different challenge: direct text rewriting can misalign tokens with their labels.  
To address this, we use a placeholder-based mechanism.
Before generation, we replace each entity span with a unique placeholder (e.g., \texttt{\_\_ENT0\_\_}) and instruct the LLM to rewrite only the surrounding context while preserving all placeholders.
After generation, we expand placeholders back to their original entity tokens with corresponding BIO tags, ensuring perfect token--label alignment.
To guide the generation, we provide $k$ reference examples---other high-loss samples from the same slice with the same label---with entities masked as type markers (e.g., \texttt{<ENT:PER>}) to prevent direct copying while allowing the model to borrow stylistic patterns.

To ensure quality, we generate $m$ candidates per seed and apply a two-stage filtering pipeline. 
First, we apply hard constraints to verify format compliance, including length limits and single-line output; for NER tasks, we additionally enforce that each placeholder appears exactly once and that original entity surface forms do not appear in the rewritten context. 
Second, we score surviving candidates to balance prediction consistency and semantic diversity. 
Specifically, we pass each candidate through the current model and compute its score as:
\begin{equation*}
\resizebox{0.896\columnwidth}{!}{$
\displaystyle
\text{score}_j =
\lambda_1 \operatorname{sim}(\mathbf{p}_{\mathrm{aug}}^{(j)}, \mathbf{p}_{\mathrm{seed}})
+
\lambda_2 \bigl(1-\operatorname{sim}(\mathbf{z}_{\mathrm{aug}}^{(j)}, \mathbf{z}_{\mathrm{seed}})\bigr)
$}
\end{equation*}
, where $\mathbf{p}$ denotes predicted class probabilities from the task model and $\mathbf{z}$ denotes sentence embeddings from the pretrained BGE encoder, with both terms computed via cosine similarity. 
The first term rewards prediction consistency, while the second encourages semantic diversity relative to the seed.
We then select the top-scoring candidates up to the per-seed budget, typically retaining 2–3 augmentations per seed to balance quality and quantity.

\subsection{Iterative Training Framework}

The latent slice discovery, distribution gap analysis, and targeted augmentation mechanisms integrate seamlessly into a unified iterative training procedure summarized in Algorithm~\ref{alg:framework}.
Training proceeds through $T$ rounds, beginning with $R_w$ warmup rounds of standard training to establish initial model behavior before augmentation begins.
At each subsequent round, the framework trains the model on the current dataset, re-discovers slices using updated joint representations, identifies deficit labels within misaligned slices, and generates targeted augmentations that are merged into the training set for the next round.
Re-slicing at each round allows the discovery mechanism to adapt as the model evolves, enabling progressive correction of local distribution defects without predefined groups or manual intervention.

\begin{algorithm}[tb]
    \caption{AIA$^{2}$ unified training procedure}
    \label{alg:framework}
    \textbf{Input}: Training set $\mathcal{D}_0$, validation set $\mathcal{D}_{\text{val}}$\\
    \textbf{Parameter}: Warmup rounds $R_w$, total rounds $T$, gap threshold $\tau$\\
    \textbf{Output}: Best model $\theta^*$
    \begin{algorithmic}[1]
        \STATE Initialize $\theta$, $\mathcal{D} \gets \mathcal{D}_0$, $\theta^* \gets \theta$
        \FOR{$t = 1$ to $T$}
            \STATE $\theta \gets \textsc{Train}(\theta, \mathcal{D})$
            \STATE Update $\theta^*$ if validation on $\mathcal{D}_{\text{val}}$ improves
            \IF{$t \leq R_w$ or $t = T$}
                \STATE \textbf{continue}
            \ENDIF
            \STATE Compute $\mathbf{u}_i=[\bar{\mathbf{z}}_i \,||\, \alpha\bar{\mathbf{p}}_i]$, $\forall (x_i,y_i)\in\mathcal{D}$
            \STATE Discover slices $\{S_s\}$ using Leiden on the kNN graph
            \FOR{each slice $s$ with $\text{gap}_s > \tau$}
                \STATE Select high-loss seeds from deficit labels
                \STATE Generate augmentations $\mathcal{A}_s$ via LLM
            \ENDFOR
            \STATE $\mathcal{D} \gets \mathcal{D} \cup \bigcup_s \mathcal{A}_s$
        \ENDFOR
        \STATE \textbf{return} $\theta^*$
    \end{algorithmic}
\end{algorithm}

\section{Experiment}

\subsection{Data}

We evaluate our approach on five publicly available corpora across two tasks: text classification (CLS) and named-entity recognition (NER). 
These datasets exhibit both global class imbalance and heterogeneous subpopulations whose local label distributions may differ from the global pattern. 
For CLS, HateXplain~\cite{mathew2021hatexplain} provides hate speech annotations with target-community metadata; HumAID~\cite{Firoj_2020_HumAID} contains humanitarian tweets from multiple disaster sources; and WWW2015~\cite{Hovy_2015_User} includes business reviews with company-level metadata. 
For NER, RE3D~\cite{dstl_2026_re3d} covers news, government, and encyclopedic documents annotated with defense and security entities, while CrossNER~\cite{liu2020crossner} spans five specialized domains with domain-specific entity labels. 
We quantify global imbalance by $\mathrm{IR} = |\mathrm{maj}|/|\mathrm{min}|$.
Table~\ref{tab:dataset_overview} summarizes the dataset characteristics, with further details provided in Appendix~\ref{sec:appendix_data}.

\begin{table}[t]
\centering
\footnotesize
\renewcommand{\arraystretch}{1.08}
\setlength{\tabcolsep}{3.4pt}

\resizebox{\columnwidth}{!}{%
\begin{tabular}{l l c r c c r r r c}
\toprule
\multirow{2}{*}{\textbf{Dataset}} &
\multirow{2}{*}{\textbf{Domain}}  &
\multirow{2}{*}{\textbf{Task}}    &
\multirow{2}{*}{\textbf{Size}}    &
\multirow{2}{*}{\textbf{\#Classes}} &
\multirow{2}{*}{\textbf{\#Subgroups}} &
\multicolumn{3}{c}{\textbf{Splits}} &
\multirow{2}{*}{\textbf{IR}} \\
\cmidrule(lr){7-9}
& & & & & &
\textbf{Train} & \textbf{Valid} & \textbf{Test} & \\
\midrule
HateXplain & Social      & CLS & 14,995 & 3 & 8  & 11,996 & 1,499 & 1,500 & 1.6 \\
HumAID     & Crisis      & CLS & 15,311 & 7 & 5  & 10,717 & 1,559 & 3,035 & 14.8 \\
WWW2015    & Review      & CLS & 15,449 & 5 & 10 & 12,378 & 1,562 & 1,509 & 27.4 \\
RE3D       & Security    & NER & 931    & 5 & 7  & 634    & 50    & 247   & 16.3 \\
CrossNER   & Multi-domain & NER & 5,327  & 5 & 5  & 4,261  & 532   & 534   & 2.0 \\
\bottomrule
\end{tabular}%
}

\caption{Dataset characteristics. \#Subgroups indicates annotated metadata (e.g., country, target group) that define natural data partitions. IR (imbalance ratio) is the ratio of the largest class size to the smallest class size.}
\label{tab:dataset_overview}
\end{table}

\subsection{Baselines}

We compare AIA$^{2}$ against baselines spanning three paradigms: sample-level reweighting, group-level optimization, and LLM-based augmentation.
For sample-level methods, \textit{Focal Loss}~\cite{lin2017focal} down-weights well-classified examples using the modulating factor $(1-p_t)^\gamma$, while \textit{Just Train Twice (JTT)}~\cite{Just2021Liu} upweights examples misclassified by an early ERM model.
For group-level methods, \textit{Group DRO}~\cite{Distributionally_2020_Shiori} optimizes the worst-group loss through group reweighting, and \textit{Diverse Prototypical Ensembles (DPE)}~\cite{to2025diverse} uses subgroup annotations to build attribute-balanced validation sets and train diverse prototypical classifiers.
\textit{GEORGE}~\cite{sohoni2020no} infers hidden subclasses by clustering learned representations and applies group-robust training over the resulting proxy groups.
Following its original example-level formulation, we evaluate GEORGE only on the three classification datasets.
AugGPT paraphrases random training samples; CB-LLM augments minority classes by the global class distribution.
Unlike these baselines, AIA$^{2}$ performs targeted augmentation by inferring latent subgroup-label structure.
Additional baseline details are provided in Appendix~\ref{sec:appendix_baseline}.

\subsection{Experimental Settings}

To ensure consistency with the baselines, we use DeBERTa-v3-base~\cite{he2021debertav3} as the default task model.
For AIA$^{2}$, we use BGE-large-en-v1.5~\cite{bge_embedding} for slice discovery and Qwen3-32B~\cite{qwen3technicalreport} for sample augmentation (temperature 0.85).
We run 8 iterative rounds and select the best checkpoint based on validation performance.
We report Micro-F1 and Macro-F1.
For NER, scores are computed at the entity level under the BIO scheme, excluding the O tag.
We further evaluate worst-group performance using WGA for multi-class classification and WG-F1 for NER.
WGA is the minimum accuracy across predefined subgroups, while WG-F1 is the minimum micro-F1 across subgroups.
All results are averaged over five independent runs with different random seeds, and Appendix~\ref{sec:appendix_std} reports standard deviations of the results in Table~\ref{tab:cls_ner_results_wg}.
Additional implementation details, hyperparameter sensitivity analyses, and computational overhead appear in Appendix~\ref{sec:appendix_implementation}, Appendix~\ref{sec:hyperparameter_sensitivity}, and Appendix~\ref{sec:computational}, respectively.

\begin{table*}[t]
\centering
\footnotesize
\renewcommand{\arraystretch}{1.2}
\setlength{\tabcolsep}{3pt}

\resizebox{0.82\textwidth}{!}{%
\begin{tabular}{l *{15}{c}}
\toprule
\multirow{3}{*}{\textbf{Method}}
& \multicolumn{9}{c}{\textbf{CLS}}
& \multicolumn{6}{c}{\textbf{NER}} \\
\cmidrule(lr){2-10}\cmidrule(lr){11-16}
& \multicolumn{3}{c}{\textbf{HumAID}}
& \multicolumn{3}{c}{\textbf{WWW2015}}
& \multicolumn{3}{c}{\textbf{HateXplain}}
& \multicolumn{3}{c}{\textbf{RE3D}}
& \multicolumn{3}{c}{\textbf{CrossNER}} \\
\cmidrule(lr){2-4}\cmidrule(lr){5-7}\cmidrule(lr){8-10}\cmidrule(lr){11-13}\cmidrule(lr){14-16}
& \multicolumn{1}{c}{M-F1} & \multicolumn{1}{c}{$\mu$-F1} & \multicolumn{1}{c}{WGA}
& \multicolumn{1}{c}{M-F1} & \multicolumn{1}{c}{$\mu$-F1} & \multicolumn{1}{c}{WGA}
& \multicolumn{1}{c}{M-F1} & \multicolumn{1}{c}{$\mu$-F1} & \multicolumn{1}{c}{WGA}
& \multicolumn{1}{c}{M-F1} & \multicolumn{1}{c}{$\mu$-F1} & \multicolumn{1}{c}{WG-F1}
& \multicolumn{1}{c}{M-F1} & \multicolumn{1}{c}{$\mu$-F1} & \multicolumn{1}{c}{WG-F1} \\
\midrule
Focal            & 67.5 & 76.0 & 69.5 & 54.2 & 85.2 & 78.4 & 63.3 & 65.1 & 54.3 & 64.6 & 64.0 & 59.1 & 71.4 & 72.6 & 60.9 \\
GroupDRO         & 66.8 & 75.9 & 71.5 & 53.1 & 85.4 & 76.7 & 63.1 & 64.8 & 54.9 & 64.9 & 63.9 & 60.0 & 73.7 & 72.0 & 61.9 \\
GEORGE       & 67.2 & 75.9 & 70.4  & 56.2 & 84.5 & 78.8  & 63.4 & 64.4 & 56.4 & -- & -- & -- & -- & -- & -- \\
JTT              & 65.6 & 74.5 & 68.7 & 52.6 & 86.1 & 79.4 & 59.3 & 61.7 & 52.0 & 64.4 & 62.3 & 57.8 & 72.2 & 73.4 & 62.5 \\
DPE              & 53.3 & 75.4 & 69.6 & 43.4 & 86.4 & 79.4 & 62.7 & 63.8 & 56.7 & 63.5 & 63.2 & 60.1 & 72.2 & 73.4 & 61.9 \\
CB-LLM           & 66.4 & 75.6 & 67.5 & 53.5 & 85.1 & 78.8 & 62.0 & 64.1 & 55.6 & 64.7 & 63.7 & 59.5 & 71.7 & 72.3 & 60.9 \\
AugGPT           & 67.5 & 76.0 & 71.0 & 54.4 & 85.5 & 77.0 & 60.4 & 63.3 & 53.7 & 64.9 & 64.1 & 59.6 & 72.1 & 72.5 & 62.2 \\
AIA$^{2}$ (Ours)  & \textbf{69.4} & \textbf{77.6} & \textbf{71.8} & \textbf{57.6} & \textbf{87.5} & \textbf{80.9} & \textbf{64.6} & \textbf{66.2} & \textbf{57.7} & \textbf{66.5} & \textbf{66.3} & \textbf{60.5} & \textbf{74.4} & \textbf{74.9} & \textbf{63.8} \\
\bottomrule
\end{tabular}%
}
\caption{Results of baselines and the proposed AIA$^{2}$ (percentages).
M-F1 and $\mu$-F1 denote macro-averaged and micro-averaged F1 scores, respectively. 
WGA denotes worst-group accuracy for CLS, and WG-F1 denotes worst-group F1 score for NER. 
GEORGE is evaluated only on the classification datasets.
Best scores are in \textbf{bold}.}

\label{tab:cls_ner_results_wg}
\end{table*}

\begin{figure*}[hbt]
\centering
\includegraphics[width=1\textwidth]{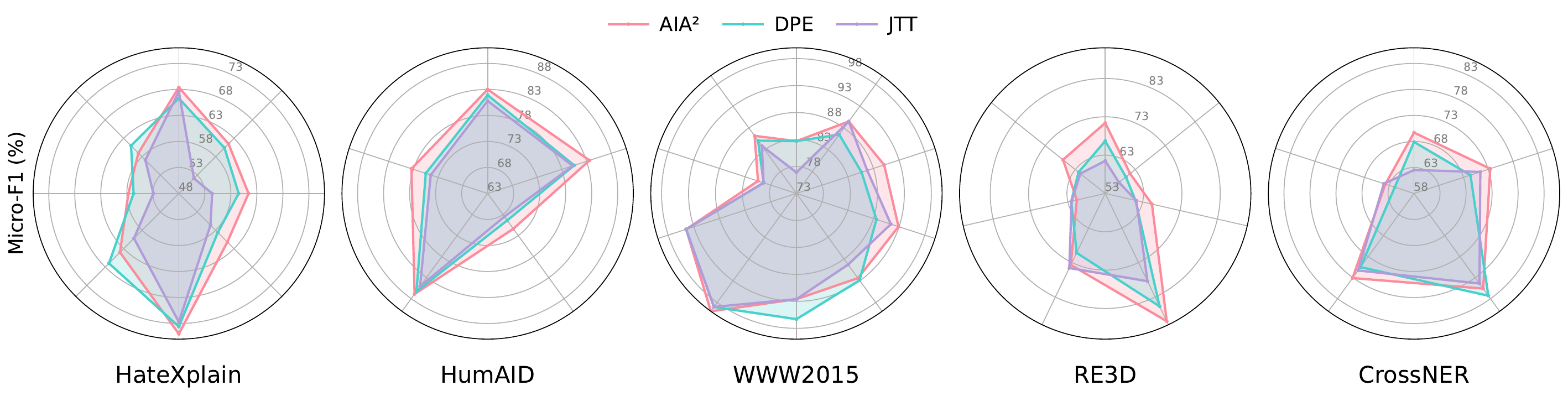} 
\vspace{-4pt}
\caption{
Subgroup-level Micro-F1 (\%) across five datasets.
Each radar plot compares AIA$^{2}$ with two representative baselines, JTT and DPE.
Each spoke denotes a subgroup, and radial values indicate Micro-F1.
}
\label{result_subgroup}
\end{figure*}

\section{Results}

\subsection{Overall Performance}

Table~\ref{tab:cls_ner_results_wg} summarizes the overall performance of AIA$^{2}$ against seven baselines, with GEORGE evaluated on the classification datasets only.
Across datasets, AIA$^{2}$ consistently achieves the best results, improving Macro-F1, Micro-F1, and worst-group metrics by up to 1.9, 2.2, and 1.5 points, respectively.
On the severely imbalanced WWW2015 (IR = 27.4), AIA$^{2}$ reaches 57.6 Macro-F1, outperforming GEORGE by 1.4 points, while improving WGA over the best baseline by 1.5 points.
It also remains effective in less skewed settings, improving both Macro-F1 and WG-F1 on CrossNER (IR = 2.0).
The comparisons with AugGPT and CB-LLM show that these gains are not simply due to generic LLM paraphrasing or global class-balanced expansion.
On WWW2015, AIA$^{2}$ improves over the best augmentation baseline by 3.2 points in Macro-F1, 2.0 points in Micro-F1, and 2.1 points in WGA.
Compared with reweighting, group-aware optimization, error-based upweighting or ensembling, and LLM-based augmentation, AIA$^{2}$ provides more consistent improvements.
These results suggest that discovering latent slices and repairing slice-level label deficits through targeted augmentation is more effective than relying on predefined groups, loss-based signals, or untargeted LLM augmentation.

\paragraph{Subgroup-level analysis.}
Figure~\ref{result_subgroup} presents subgroup Micro-F1 comparisons using radar plots, contrasting AIA$^{2}$ with two representative baselines, JTT and DPE.
Each radar chart corresponds to one dataset, where each spoke denotes a subgroup and the radial value indicates Micro-F1.
We choose JTT and DPE because they provide strong worst-group performance while reflecting two distinct supervision settings: JTT does not rely on subgroup labels, whereas DPE leverages group annotations.
Across all five datasets, AIA$^{2}$ forms the outer envelope on most spokes, achieving the highest or near-highest Micro-F1 for the majority of subgroups.
The gains are most evident on challenging subgroups where baselines exhibit sharp drops (e.g., several minority subgroups in HateXplain, specific disaster events in HumAID, and domain-specific subsets in RE3D/CrossNER).
Overall, the radar plots indicate that AIA$^{2}$ delivers broad and consistent improvements across diverse subgroups, rather than increasing aggregate scores by trading off performance on specific subgroups.

\begin{table*}[t]
\centering
\footnotesize
\renewcommand{\arraystretch}{1.2}
\setlength{\tabcolsep}{3.0pt}

\resizebox{0.86\textwidth}{!}{%
\begin{tabular}{c l *{15}{c}}
\toprule
\multirow{3}{*}{\textbf{Blk}}
& \multirow{3}{*}{\textbf{Method}}
& \multicolumn{9}{c}{\textbf{CLS}}
& \multicolumn{6}{c}{\textbf{NER}} \\
\cmidrule(lr){3-11}\cmidrule(lr){12-17}
& 
& \multicolumn{3}{c}{\textbf{HumAID}}
& \multicolumn{3}{c}{\textbf{WWW2015}}
& \multicolumn{3}{c}{\textbf{HateXplain}}
& \multicolumn{3}{c}{\textbf{RE3D}}
& \multicolumn{3}{c}{\textbf{CrossNER}} \\
\cmidrule(lr){3-5}\cmidrule(lr){6-8}\cmidrule(lr){9-11}\cmidrule(lr){12-14}\cmidrule(lr){15-17}
& 
& \multicolumn{1}{c}{M-F1} & \multicolumn{1}{c}{$\mu$-F1} & \multicolumn{1}{c}{WGA}
& \multicolumn{1}{c}{M-F1} & \multicolumn{1}{c}{$\mu$-F1} & \multicolumn{1}{c}{WGA}
& \multicolumn{1}{c}{M-F1} & \multicolumn{1}{c}{$\mu$-F1} & \multicolumn{1}{c}{WGA}
& \multicolumn{1}{c}{M-F1} & \multicolumn{1}{c}{$\mu$-F1} & \multicolumn{1}{c}{WG-F1}
& \multicolumn{1}{c}{M-F1} & \multicolumn{1}{c}{$\mu$-F1} & \multicolumn{1}{c}{WG-F1} \\
\midrule

\multirow{4}{*}{\textbf{MA}}
& Full & \textbf{69.4} & \textbf{77.6} & \textbf{71.8} & \textbf{57.6} & \textbf{87.5} & \textbf{80.9} & \textbf{64.6} & \textbf{66.2} & \textbf{57.7} & \textbf{66.5} & \textbf{66.3} & \textbf{60.5} & \textbf{74.4} & \textbf{74.9} & 63.8 \\
& w/o slice    & 67.8 & 76.8 & 70.8 & 54.3 & 86.5 & 80.0 & 62.3 & 64.1 & 55.2 & 63.9 & 63.9 & 58.7 & 73.2 & 71.7 & 61.1 \\
& w/o gap-aware& 67.2 & 76.1 & 69.5 & 55.5 & 86.3 & 78.9 & 62.6 & 64.5 & 57.6 & 65.0 & 64.7 & 53.3 & 73.1 & 73.1 & 62.2 \\
& w/o LLM      & 67.7 & 76.7 & 70.4 & 53.8 & 86.5 & 78.9 & 63.5 & 65.5 & 56.7 & 65.6 & 65.4 & 58.2 & 72.8 & 71.8 & \textbf{64.2} \\
\midrule

\multirow{8}{*}{\textbf{TM}}
& Focal            & 66.2 & 73.8 & 68.7 & 53.3 & 85.8 & 78.2 & 62.6 & 64.2 & 54.8 & 49.9 & 47.8 & 33.3 & 65.8 & 64.3 & 53.8 \\
& GroupDRO         & \textbf{67.2} & 75.8 & 69.5 & 52.2 & 85.5 & 77.8 & 60.4 & 61.7 & 52.9 & 49.5 & 46.8 & 27.0 & 66.2 & 65.2 & 54.5 \\
& GEORGE       & 66.5 & 75.1 & 68.7  
& 52.4 & 84.0 & 77.7  
& 62.0 & 63.5 & 53.2 & -- & -- & -- & -- & -- & -- \\
& JTT              & 66.2 & 75.8 & 70.2 & 53.8 & 85.9 & 77.1 & 62.2 & 63.7 & 53.9 & 50.7 & \textbf{51.7} & 43.3 & 65.7 & 64.8 & 54.1 \\
& DPE              & 40.4 & 67.9 & 59.7 & 17.9 & 81.0 & 67.7 & 44.3 & 50.9 & 34.3 & 45.0 & 43.7 & 28.6 & 24.7 & 29.9 & 16.1 \\
& CB-LLM           & 66.0 & 75.6 & 68.5 & 52.5 & 85.2 & 77.5 & 62.1 & 63.7 & 52.9 & 51.0 & 50.2 & 40.5 & 64.4 & 64.7 & 55.0 \\
& AugGPT           & 65.5 & 75.4 & 68.1 & 51.9 & 84.6 & 77.6 & 62.3 & 63.9 & 51.0 & 51.8 & 50.2 & 42.7 & 63.9 & 64.1 & 57.3 \\
& AIA$^{2}$ (Ours) & 67.0 & \textbf{76.1} & \textbf{69.5} & \textbf{54.9} & \textbf{86.4} & \textbf{78.2} & \textbf{63.2} & \textbf{64.5} & \textbf{54.8} & \textbf{53.8} & 51.2 & \textbf{43.6} & \textbf{66.8} & \textbf{66.1} & \textbf{59.1} \\
\midrule

\multirow{3}{*}{\textbf{BB}}
& Qwen3-32B & 69.4 & 77.6 & \textbf{71.8} & \textbf{57.6} & \textbf{87.5} & 80.9 & \textbf{64.6} & \textbf{66.2} & 57.7 & 66.5 & 66.3 & 60.5 & \textbf{74.4} & \textbf{74.9} & \textbf{63.8} \\
& Qwen3-14B & 69.3 & 77.6 & \textbf{71.8} & 54.0 & 86.6 & 80.1 & 63.3 & 65.2 & 57.7 & 64.8 & 65.0 & 60.2 & \textbf{74.4} & 73.3 & 61.5 \\
& Phi-4     & \textbf{69.5} & \textbf{77.7} & 71.7 & 55.6 & 87.2 & \textbf{81.1} & 63.6 & 65.2 & \textbf{58.9} & \textbf{68.9} & \textbf{68.0} & \textbf{62.1} & 74.3 & 72.9 & 62.8 \\

\bottomrule
\end{tabular}%
}

\caption{\textbf{Combined ablation results.}
Blocks correspond to: \textbf{MA} (module ablation of AIA$^{2}$),
\textbf{TM} (task model ablation on \textbf{ModernBERT-base}),
and \textbf{BB} (LLM backbone ablation).
In \textbf{MA}, \textbf{w/o slice} removes latent slice discovery,
\textbf{w/o gap-aware} removes label-gap-aware seed selection,
and \textbf{w/o LLM} removes LLM-based augmentation.
}
\label{tab:combined_ablation_results}
\end{table*}

\begin{table}[t]
\centering
\footnotesize
\renewcommand{\arraystretch}{1.08}
\setlength{\tabcolsep}{4.5pt}
\begin{tabular}{l c c c}
\toprule
\textbf{Dataset} & \textbf{\#Groups} & \textbf{\#Communities} & \textbf{AMI} $\uparrow$ \\
\midrule
HumAID   & 5  & 10--12 & 0.637 \\
WWW2015  & 10 & 13--14 & 0.615 \\
RE3D     & 7  & 8--10  & 0.268 \\
CrossNER & 5  & 12--15 & 0.559 \\
\bottomrule
\end{tabular}
\caption{
Alignment between discovered communities and metadata-defined groups.
\#Groups denotes the number of attribute values, and \#Communities reports the observed range across discovery rounds.
}
\label{tab:slice_attribute_alignment}
\end{table}



\subsection{Slice Alignment and Deficit Repair}
\label{sec:slice_deficit_analysis}

\paragraph{Slice--attribute alignment.}
To examine how the latent slices discovered without subgroup supervision relate to available dataset attributes, we compare their community assignments with metadata labels withheld from training and used only for post-hoc evaluation.
For the four datasets in which each instance has a single metadata value, we compute Adjusted Mutual Information (AMI), a chance-adjusted measure of agreement between partitions~\cite{vinh2009information}, over the original training instances after each community-discovery round and report the mean across rounds.
HateXplain is excluded because an instance may be associated with multiple target groups, so its metadata does not define a single partition.

As shown in Table~\ref{tab:slice_attribute_alignment}, alignment varies across datasets.
AMI reaches 0.637 on HumAID, 0.615 on WWW2015, and 0.559 on CrossNER, but decreases to 0.268 on RE3D.
Across all four datasets, the number of discovered communities exceeds the number of metadata-defined groups, suggesting that a single group may contain multiple regions with distinct semantic content or model responses.
Despite its weaker alignment, AIA$^{2}$ still improves all three evaluation metrics on RE3D in Table~\ref{tab:cls_ner_results_wg}, indicating that useful slices need not closely reproduce the available metadata partition.
Overall, the discovered slices capture part of the metadata-defined structure while also reflecting finer-grained semantic--predictive variation.

\begin{figure}[t]
\centering
\includegraphics[width=1\columnwidth]{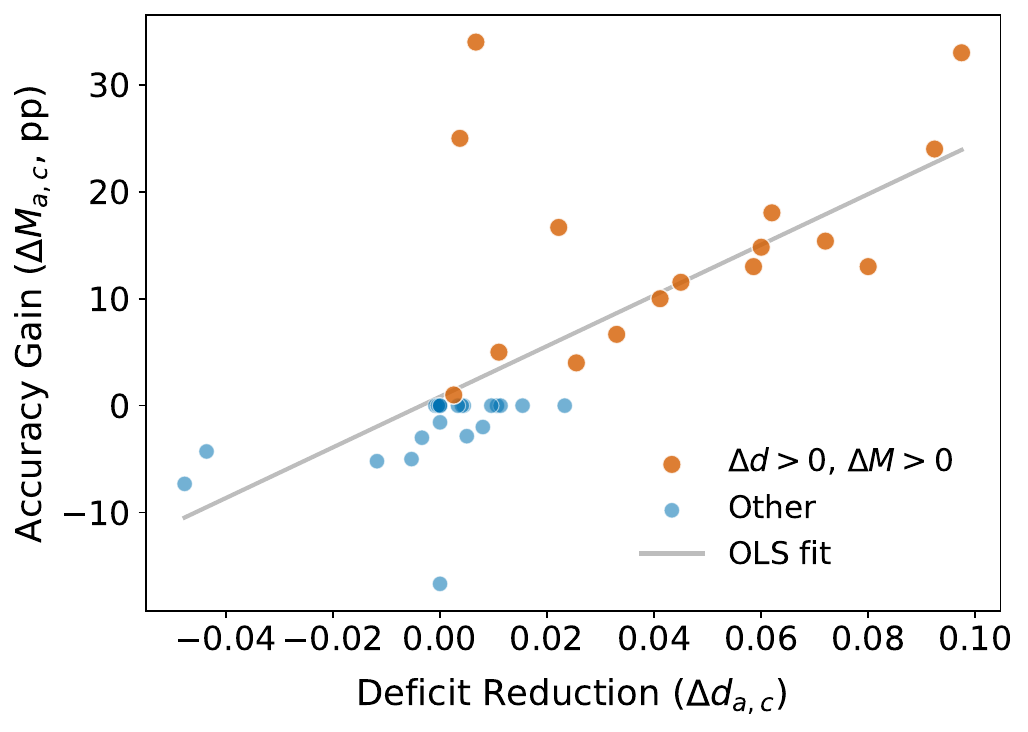}
\caption{
Subgroup-label accuracy gain versus deficit reduction on WWW2015, with an OLS fit.
}
\label{fig:result_a}
\end{figure}

\paragraph{Deficit repair and subgroup gains.}

To further examine whether AIA$^{2}$ improves subgroup robustness through targeted deficit repair, we conduct a subset-level diagnostic analysis on classification datasets, where subgroup-label subsets and subset-level accuracy are directly defined.
A subgroup-label subset $(a,c)$ refers to examples of label $c$ within subgroup $a$.
For each subset, its label-deficit gap measures how much label $c$ is underrepresented in subgroup $a$ compared with the global label distribution of the original training set.
We define the deficit reduction after augmentation as
$\Delta d_{a,c}=d_{a,c}^{\mathrm{pre}}-d_{a,c}^{\mathrm{post}}$,
where a larger value indicates that augmentation better reduces the local label deficit.
We then compute the corresponding accuracy gain as
$\Delta M_{a,c}=M_{a,c}^{\mathrm{AIA}}-M_{a,c}^{\mathrm{NoAug}}$,
where $M_{a,c}^{\mathrm{NoAug}}$ is obtained from the same training setup without generated data.

On WWW2015, deficit reduction is strongly associated with accuracy gain across 39 subgroup-label subsets: the Spearman correlation is 0.769 ($p=1.1\times10^{-8}$), and the ordinary least squares (OLS) fit explains a substantial portion of the variation ($R^2=0.487$).
Figure~\ref{fig:result_a} visualizes this pattern, showing that larger $\Delta d_{a,c}$ generally corresponds to larger $\Delta M_{a,c}$.
Among all subsets, 16 subsets (41.0\%) fall into the positive-repair region with both reduced deficit and improved accuracy.
This result suggests that AIA$^{2}$ gains are concentrated in subsets where subgroup-label deficits are effectively reduced, supporting targeted deficit repair as a key mechanism behind the observed subgroup robustness improvements.
Additional results on the other two CLS datasets are provided in Appendix~\ref{sec:additional_generated_analysis} and show the same positive trend.

\subsection{Ablation Study: Module Contributions}

We conduct a module-level ablation study by selectively disabling one module at a time while keeping the rest of the iterative training pipeline unchanged:
Latent Slice Discovery (``w/o slice''), Label-Gap-Aware Seed Selection (``w/o gap-aware''), and LLM-Based Augmentation (``w/o LLM'').
In the first ablation, learned slices are replaced with random partitions of equal cardinality.
In the second, seed examples are sampled uniformly at random within each slice rather than prioritized by type-specific gap scores.
In the third, the LLM generator is replaced with a copy-based method that simply duplicates the seed instances.

As shown in the first block (\textbf{MA}) of Table~\ref{tab:combined_ablation_results}, the full model performs best across benchmarks, achieving the top results on nearly all metrics.
Removing Latent Slice Discovery leads to consistent degradation, with drops on WWW2015 Macro-F1 (57.6 to 54.3) and CrossNER Micro-F1 (74.9 to 71.7), indicating that learned slices support targeted augmentation.
Removing Label-Gap-Aware Seed Selection weakens tail performance, with the largest drop on RE3D WG-F1 (60.5 to 53.3), showing that gap-based seed prioritization is critical for under-performing labels.
Replacing LLM-Based Augmentation with simple duplication reduces most scores, especially on WWW2015 Macro-F1 (57.6 to 53.8), confirming that diverse examples provide richer supervision than duplicates.

\subsection{Ablation Study: Impacts of Base Models}

We investigate the robustness of AIA$^{2}$ to two key model choices:
1) the task backbone used for classification and NER; 
2) the LLM used for data augmentation.
Additional ablations of slice representation, seed selection, and clustering methods are provided in Appendix~\ref{app:additional_ablations}.

\paragraph{Task model generalization.}
The \textbf{TM} block of Table~\ref{tab:combined_ablation_results} evaluates whether AIA$^{2}$ remains effective when the task model is changed from DeBERTa-v3-base to ModernBERT-base~\cite{modernbert}.
Although all methods show slightly lower absolute performance under ModernBERT, AIA$^{2}$ still achieves the strongest or comparable results on most benchmarks.
In particular, it obtains the best Macro-F1 on four of the five datasets and the best worst-group score on both NER datasets.
For example, on CrossNER, AIA$^{2}$ improves WG-F1 over the best baseline (AugGPT, 57.3) by 1.8 points.
We also find that DPE is more sensitive to the backbone change, likely because it relies on frozen ERM features, whereas AIA$^{2}$ remains comparatively stable across encoder architectures.
These results suggest that slice-aware augmentation is not tied to a specific task backbone and generalizes across modern pretrained encoders.

\paragraph{Data Generator Impacts.}
The \textbf{BB} block of Table~\ref{tab:combined_ablation_results} examines sensitivity to augmentation LLM choice by comparing Qwen3-32B (default), Qwen3-14B, and Phi-4.
Overall, results are stable across generators, with most datasets varying within 1--2 Macro-F1 points (e.g., HumAID: 69.3--69.5).
Phi-4 performs particularly well on RE3D (Macro-F1 68.9, WG-F1 62.1), exceeding Qwen3-32B by 2.4 and 1.6 points, suggesting that smaller but well-aligned models can be effective for targeted augmentation.
The main sensitivity occurs on WWW2015, where Qwen3-32B (57.6) outperforms Qwen3-14B (54.0) by 3.6 Macro-F1 points, likely due to the higher diversity required for business review generation.
These findings indicate that AIA$^{2}$ is not tightly coupled to a specific LLM and can leverage a range of open-source generators without substantial performance loss.

\section{Related Work}

Class imbalance and group robustness have been widely studied in machine learning. 
Re-weighting methods adjust loss contributions by class frequency or training difficulty, as in Focal Loss~\cite{lin2017focal} and Class-Balanced Loss~\cite{cui2019class}. 
Group-robust methods further target performance disparities by minimizing worst-case loss over predefined groups~\cite{Distributionally_2020_Shiori}. 
Later methods reduce the need for full group annotations by using initial model errors, biased models, or inferred environments to identify hard examples~\cite{Just2021Liu,nam2020learning,creager2021environment}. 
However, these methods usually rely on observed group information during training or validation, or address global class imbalance without modeling latent subgroup structure.

Another line of work discovers underperforming subgroups without explicit annotations. 
Domino identifies coherent error slices using cross-modal embeddings and produces natural language descriptions~\cite{eyuboglu2022domino}. 
GEORGE clusters learned representations to uncover hidden subclasses, then applies group-robust training with pseudo-labels~\cite{sohoni2020no}. 
Spotlight searches representation space for regions with concentrated errors~\cite{d2022spotlight}. 
These methods are useful for diagnosis, but they mainly identify failure modes after training. 
They do not integrate subgroup discovery into training to repair local distributional gaps.

Data augmentation offers a complementary way to improve robustness. 
Traditional text augmentation uses word-level perturbations or back-translation to create paraphrases~\cite{wei2019eda,sennrich2016improving}. 
Recent LLM-based methods generate more fluent augmented examples through sample mixing, rephrasing, or counterfactual prompting~\cite{han-etal-2025-attributes,yoo2021gpt3mix,dai2025auggpt,chen2023disco}. 
Mixup-based methods interpolate between training examples, while LISA selectively mixes samples using domain or label information to improve out-of-distribution robustness~\cite{guo2019augmenting,yao2022improving}. 
However, existing augmentation strategies usually apply transformations globally or depend on explicit domain annotations for selection. 
They lack a way to locate distributionally deficient latent subgroups and direct augmentation toward those regions.

\section{Conclusion}

This paper presented AIA$^{2}$, a framework for improving subgroup robustness without explicit attribute annotations. 
AIA$^{2}$ discovers latent slices in a joint semantic-predictive space, identifies local label deficits, and uses LLMs for targeted augmentation. 
Across five text classification and named entity recognition datasets, AIA$^{2}$ improves overall and worst-group performance over competitive baselines. 
Generated-data analysis shows that gains concentrate in subgroup-label subsets where augmentation reduces local deficits, supporting targeted deficit repair as the main mechanism. 
Ablation studies confirm the roles of slice discovery, gap-aware selection, and LLM-based generation, with stable results across design choices.
Future work will explore more efficient slice discovery and augmentation strategies to reduce the computational overhead of iterative training. It will also be benefit to extend AIA$^{2}$ to broader NLP tasks and systematically evaluate its effectiveness across subgroups defined by demographic attributes, such as health informatics, where rich demographic attributes exist and imbalance and augmentation may interact with demographic biases.

\section*{Limitations}

Despite its effectiveness, AIA$^{2}$ has several limitations. 
First, the framework introduces extra computational cost due to iterative training, repeated slice discovery, distribution gap analysis, and LLM-based augmentation. 
Although the overhead is manageable in our experiments, scaling to larger corpora or stronger task models may require more efficient slice construction and generation strategies. 
Second, latent slice discovery depends on the quality of semantic embeddings and prediction signals. 
When these representations are noisy or weakly aligned with the target domain, the discovered slices may become less reliable, which can affect seed selection and augmentation quality. 
Third, LLM-generated examples may contain label noise, hallucinated details, or demographic and topical biases. Although our filtering procedure checks format, label consistency, and semantic diversity, it cannot fully eliminate these risks.
Finally, our evaluation focuses on text classification and named entity recognition. 
The generalizability of AIA$^{2}$ to other NLP tasks, such as relation extraction or question answering, remains to be further validated. 
Future work can improve the efficiency, stability, and applicability of slice-aware augmentation.

\section*{Acknowledgments}
The first and second authors were supported by the NSF CRII (IIS-2245920) and MRI (CNS-2318210) awards, respectively.
We thank the computing resources provided by the iTiger GPU cluster~\cite{itiger} supported by the NSF MRI program under the award CNS-2318210.

\bibliography{custom}

\appendix

\section{Data Imbalance Details}
\label{sec:appendix_data}

\begin{figure*}[hbt]
\centering
\includegraphics[width=1\textwidth]{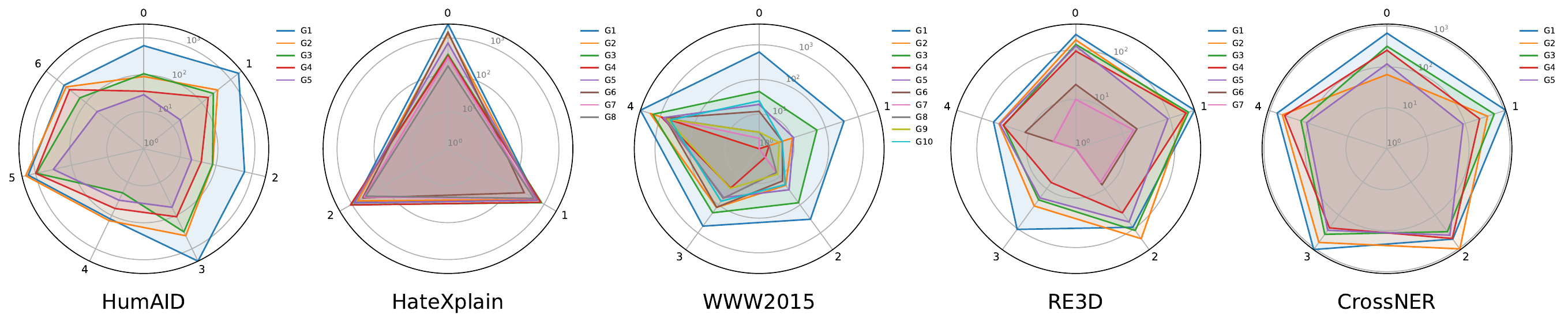} 
\caption{Label distributions across subgroups for five datasets. 
Each polygon represents a subgroup (G1, G2, ...), and axes correspond to label indices. 
Values are shown on a logarithmic scale ($\log_{10}$).
}
\label{five_radar}
\end{figure*}

\paragraph{HumAID}\cite{Firoj_2020_HumAID} contains 15,311 annotated tweets collected from five different natural disasters (e.g., Hurricane Irma 2017, Cyclone Idai 2019). 
These disaster events serve as natural subgroups, with distinct label distributions across humanitarian categories as shown in Figure~\ref{five_radar}. 
The tweets are labeled into humanitarian categories (e.g., Infrastructure and utility damage, Rescue volunteering or donation effort). 
The moderate class imbalance (IR = 14.8) and social media context test a model's robustness to noisy, informal text with uneven class distributions.

\paragraph{HateXplain}\cite{mathew2021hatexplain} contains 14,995 annotated posts collected from Twitter and Gab across eight target community groups (e.g., African, Caucasian, Homosexual).
These target communities serve as natural subgroups, with distinct label distributions across hate speech categories as shown in Figure~\ref{five_radar}.
The posts are labeled into three categories: hatespeech, offensive, and normal.
The relatively balanced class distribution (IR = 1.6) and social media context test a model's ability to distinguish fine-grained boundaries between hate speech and offensive language in informal, user-generated text.

\paragraph{WWW2015}~\cite{Hovy_2015_User} contains 15,449 business reviews collected from ten different companies.
These company sources serve as natural subgroups, with distinct label distributions across rating levels as shown in Figure~\ref{five_radar}.
The reviews are labeled into five rating categories (1-5 stars).
The substantial class imbalance (IR = 27.4) and review domain test a model's robustness to skewed sentiment distributions and domain-specific language patterns across different business contexts.

\paragraph{RE3D}~\cite{dstl_2026_re3d} contains 931 annotated documents collected from seven different sources spanning news, government, and encyclopedic domains (e.g., BBC\_Online, CENTCOM, UK\_Government).
These document sources serve as natural subgroups, with distinct entity distributions across defense and security contexts as shown in Figure~\ref{five_radar}.
The documents are annotated with five entity types (e.g., Temporal, Organisation).
The moderate class imbalance (IR = 16.3) and security domain test a model's ability to recognize specialized entities in formal, domain-specific text with limited training data.

\paragraph{CrossNER}~\cite{liu2020crossner} contains 5,327 annotated sentences collected from five specialized domains (Politics, Natural Science, Music, Literature, and Artificial Intelligence).
These domains serve as natural subgroups, with distinct entity distributions across knowledge areas as shown in Figure~\ref{five_radar}.
The sentences are annotated with five entity types (e.g., Location, Quantity, Temporal, Organisation, Person).
The relatively balanced class distribution (IR = 2.0) and multi-domain setting test a model's robustness to domain shift and its ability to generalize entity recognition across diverse topical contexts.

\section{Baseline Details}
\label{sec:appendix_baseline}

\paragraph{Focal Loss}~\cite{lin2017focal} adds a modulating factor to focus on data samples with hard minority classes and down-weight well-classified (easy) instances. The resulting loss is formulated as $\mathcal{L}_{\text{Focal}} = -\alpha(1 - p_t)^{\gamma}\log p_t$, where $p_t$ refers to a predicted probability for the true class.
Following prior practice, we set the focusing parameter $\gamma=2$ and adopt a class-balanced weighting schema for $\alpha$.

\paragraph{Group DRO}~\cite{Distributionally_2020_Shiori} is a group-aware robust baseline that uses observed group labels and optimizes a worst-case objective $\min_{\theta}\max_{g\in\mathcal{G}}\mathcal{L}_g(\theta)$ over predefined groups. 
In practice, it maintains a distribution over groups $q$ and upweights groups with higher minibatch loss via an exponentiated-gradient update, then updates model parameters by minimizing the resulting group-reweighted loss. 
We implement Group DRO on top of the same ERM backbone and use training-set group annotations to compute per-group minibatch losses and update $q$.

\paragraph{GEORGE}~\cite{sohoni2020no} discovers latent subclasses by clustering learned representations within each class and then applies group-robust training over the inferred subclass labels.
We follow its original example-level formulation and evaluate it on the three classification datasets.

\paragraph{Just Train Twice (JTT)}~\cite{Just2021Liu} improves worst-group robustness without requiring training group labels via a two-stage reweighting scheme. 
It first trains an ERM identification model for $T$ steps and collects the examples it misclassifies into an error set $E$. 
It then trains the final model by upweighting (equivalently, upsampling) examples in $E$ by a factor $\lambda_{\mathrm{up}}$ under standard ERM. 
Hyperparameters $(T,\lambda_{\mathrm{up}})$ are selected by worst-group performance on a small group-annotated validation set.

\paragraph{Diverse Prototypical Ensembles (DPE)}~\cite{to2025diverse} improves worst-group robustness under subpopulation shift by training a diversified ensemble of distance-based prototypical classifiers on top of frozen ERM features. 
It first fits an ERM backbone, then replaces the linear head with multiple prototype heads that assign class probabilities by a softmax over distances to class prototypes in the embedding space.
To avoid redundant members, DPE regularizes similarity among prototypes across ensemble heads, encouraging complementary decision rules that better cover minority subpopulations.

\paragraph{Class-Balanced LLM Augmentation (CB-LLM)}~\citep{valizadehaslani2024two} is based on class-balanced ChatGPT augmentation for class-imbalanced text classification.
The original method first trains on class-balanced augmented data before standard fine-tuning, so minority classes receive stronger supervision early in training.
We adopt this class-balancing principle as an augmentation baseline by prioritizing classes with fewer training samples under the global class distribution.
We use Qwen3-32B to generate label-preserving samples for these selected classes, so this baseline tests whether the gain mainly comes from increasing minority-class sample counts.

\paragraph{AugGPT}~\citep{dai2025auggpt} uses ChatGPT to rephrase training samples into label-preserving variants for text data augmentation.
The method increases training data by generating multiple semantically related versions of each input sentence.
Following this idea, we build an LLM paraphrasing baseline by randomly selecting training samples and generating paraphrases with the original labels.
We use Qwen3-32B as the generation model instead of ChatGPT, so this baseline tests whether the gain comes from adding generic LLM-generated data.

\begin{table*}[t]
\centering
\footnotesize
\renewcommand{\arraystretch}{1.08}
\setlength{\tabcolsep}{3.0pt}

\resizebox{0.95\textwidth}{!}{%
\begin{tabular}{l *{15}{c}}
\toprule
\multirow{3}{*}{\textbf{Method}}
& \multicolumn{9}{c}{\textbf{CLS}}
& \multicolumn{6}{c}{\textbf{NER}} \\
\cmidrule(lr){2-10}\cmidrule(lr){11-16}
& \multicolumn{3}{c}{\textbf{HumAID}}
& \multicolumn{3}{c}{\textbf{WWW2015}}
& \multicolumn{3}{c}{\textbf{HateXplain}}
& \multicolumn{3}{c}{\textbf{RE3D}}
& \multicolumn{3}{c}{\textbf{CrossNER}} \\
\cmidrule(lr){2-4}\cmidrule(lr){5-7}\cmidrule(lr){8-10}\cmidrule(lr){11-13}\cmidrule(lr){14-16}
& \multicolumn{1}{c}{M-F1} & \multicolumn{1}{c}{$\mu$-F1} & \multicolumn{1}{c}{WGA}
& \multicolumn{1}{c}{M-F1} & \multicolumn{1}{c}{$\mu$-F1} & \multicolumn{1}{c}{WGA}
& \multicolumn{1}{c}{M-F1} & \multicolumn{1}{c}{$\mu$-F1} & \multicolumn{1}{c}{WGA}
& \multicolumn{1}{c}{M-F1} & \multicolumn{1}{c}{$\mu$-F1} & \multicolumn{1}{c}{WG-F1}
& \multicolumn{1}{c}{M-F1} & \multicolumn{1}{c}{$\mu$-F1} & \multicolumn{1}{c}{WG-F1} \\
\midrule
Focal            & 0.53 & 0.38 & 0.77 & 0.74 & 0.47 & 1.02 & 0.45 & 0.34 & 0.59 & 0.92 & 0.58 & 1.24 & 0.60 & 0.45 & 0.81 \\
GroupDRO         & 0.74 & 0.59 & 1.38 & 1.08 & 0.70 & 2.17 & 0.69 & 0.51 & 0.94 & 1.27 & 0.86 & 2.38 & 0.80 & 0.64 & 1.52 \\
GEORGE       & 0.85 & 0.53 & 1.04 & 0.72 & 0.91 & 1.10 & 0.78 & 0.61 & 0.97
& -- & -- & -- & -- & -- & -- \\
JTT              & 1.01 & 0.80 & 1.59 & 1.46 & 1.01 & 2.44 & 0.90 & 0.66 & 1.11 & 1.78 & 1.14 & 2.89 & 1.19 & 0.81 & 1.84 \\
DPE              & 0.60 & 0.45 & 0.81 & 0.80 & 0.54 & 1.11 & 0.51 & 0.34 & 0.66 & 0.96 & 0.62 & 1.40 & 0.63 & 0.50 & 0.95 \\
CB-LLM           & 0.70 & 0.52 & 1.01 & 0.98 & 0.63 & 1.46 & 0.59 & 0.42 & 0.77 & 1.14 & 0.82 & 1.68 & 0.76 & 0.57 & 1.19 \\
AugGPT           & 0.88 & 0.61 & 1.22 & 1.16 & 0.77 & 1.67 & 0.71 & 0.56 & 0.91 & 1.34 & 0.97 & 1.89 & 0.96 & 0.71 & 1.31 \\
AIA$^{2}$ (Ours)  & 0.72 & 0.53 & 1.10 & 0.99 & 0.65 & 1.24 & 0.61 & 0.47 & 0.83 & 1.02 & 0.84 & 1.94 & 0.81 & 0.56 & 1.12 \\
\bottomrule
\end{tabular}%
}

\caption{
Standard deviation ($\pm \sigma$) over 5 runs.
M-F1 and $\mu$-F1 denote macro-averaged and micro-averaged F1 scores, respectively. 
WGA denotes worst-group accuracy for CLS, and WG-F1 denotes worst-group F1 score for NER. 
GEORGE is evaluated only on the classification datasets.
}

\label{tab:cls_ner_std}
\end{table*}

\section{Implementation Details}
\label{sec:appendix_implementation}

To validate our approach, we implement all methods using HuggingFace Transformers~\cite{wolf_2020_transformers} and PyTorch~\cite{Paszke_2019_PyTorch} and conduct experiments on the same data splits. 
All experiments were conducted on a server with an NVIDIA H100 GPU. 
We fine-tuned the microsoft/deberta-v3-base\footnote{\url{https://huggingface.co/microsoft/deberta-v3-base}} model as the backbone for all classification and NER tasks. 
We conduct iterative training for 8 rounds, with 1 warmup round for initial model training.
Each round trains for 1 epoch using the AdamW optimizer with learning rate $2 \times 10^{-5}$, batch size 16, and maximum sequence length 512.
The best model is selected based on validation combined score (average of micro- and macro-F1) across all rounds.

For slice discovery, we employ the bge-large-en-v1.5 model\footnote{\url{https://huggingface.co/BAAI/bge-large-en-v1.5}} to generate semantic embeddings, which are combined with predictive representations using $\alpha=0.1$ weighting.
We construct a k-nearest neighbor graph ($k=15$) in the joint semantic-predictive space and apply Leiden clustering~\cite{traag2019leiden} with resolution 1.0 to identify latent slices.
Only slices with size $\geq 100$ samples are considered for augmentation to ensure statistical reliability. 
For data augmentation, we use Qwen3-32B~\cite{qwen3technicalreport} as our generative model, with temperature 0.85 and top-p 0.95 to balance diversity and quality in generated samples.
For each identified slice with distribution gap exceeding threshold $\tau=0.1$, we select hard examples based on their loss values and generate augmentations with the following constraints:
(i) maximum 80 augmentations per slice with 2 augmentations per seed example,
(ii) generated texts must maintain label consistency as verified by the current classifier,
(iii) semantic diversity is enforced through cosine similarity filtering.
The augmentation budget is allocated proportionally to the deficit mass with augmentation coefficient $\rho=0.2$ and size exponent $\beta=0.8$.

\section{Prompts and Filtering}
\label{app:prompts_and_filtering}

This appendix provides the generation prompts and candidate-filtering rules used in AIA$^{2}$. Fields enclosed in braces are instantiated at runtime.

\smallskip
\noindent\textbf{Classification prompt}\par

\begin{lstlisting}[backgroundcolor=\color{codebg},basicstyle=\ttfamily\footnotesize,breaklines=true,breakindent=0pt,breakautoindent=false]
You are a careful assistant that follows instructions strictly.

You are generating data augmentation for a text classification task.
Label of the seed sample: {label}

Hard constraints:
- Output ONLY one final text.
- One line only. No quotes. No prefixes. No explanations.
- Max length: {max_chars_cap} characters.

Required transformations (must do BOTH):
1) Sentence-structure transformation (change syntax/order; do NOT just swap a few words).
2) Entity substitution with same-type entities (person->person, city->city, org->org, group->group, etc.).

Notes:
- You MAY introduce new details, but the final text must still belong to label '{label}'.
- Use the reference examples to borrow entities/structures, but do NOT copy any full sentence.

Seed text:
{seed}

Reference examples (same label, high-loss; for style/entity ideas):
{refs}

Write the new text now:
\end{lstlisting}

\noindent\textbf{NER prompt}\par

\begin{lstlisting}[backgroundcolor=\color{codebg},basicstyle=\ttfamily\footnotesize,breaklines=true,breakindent=0pt,breakautoindent=false]
You follow hard constraints strictly.

You are generating data augmentation for a token-level NER task.
You will be given a tokenized sentence (tokens are already split).

Reference examples (masked entities; <ENT:TYPE> tokens appear ONLY here; do NOT output them):
{refs_block}

Hard constraints (must satisfy all):
- Output ONLY one line.
- Output must be space-separated tokens (SINGLE spaces).
- You MUST keep every placeholder token EXACTLY unchanged.
- Each placeholder token must appear EXACTLY once in the output.
- Placeholders must be standalone tokens: do NOT attach any punctuation/quotes to them (e.g., "__ENT0__ ," not "__ENT0__,").
- Entities MUST be represented ONLY by placeholders (no entity surface forms).
- Rewrite the non-placeholder context substantially; avoid copying long consecutive spans from the seed.
- Keep punctuation tokenization style (punctuation is its own token if shown).

Placeholders (types only):
{placeholders_block}

Seed template tokens (with placeholders):
{seed_tokens_block}

Output tokens now (space-separated, one line):
\end{lstlisting}

\noindent\textbf{Candidate filtering}\par

\noindent For each selected seed, we generate 15 candidates and retain at most two. For classification, we discard empty outputs, outputs longer than 280 characters, and exact duplicates, and then select candidates according to predictive consistency with the seed and semantic diversity. For NER, we discard candidates that do not preserve every placeholder exactly once, contain reference markers or original entity surface forms, duplicate another candidate, or share at least 60\% of their placeholder-stripped 5-grams with the seed; among the remaining candidates, we prioritize semantic diversity and use predictive similarity to break ties.

\section{Stability Analysis}
\label{sec:appendix_std}

To assess run-to-run stability, we repeat the experiments in Table~\ref{tab:cls_ner_results_wg} five times with different random seeds (42, 43, 44, 45, 46) and report the standard deviation ($\pm\sigma$) of all metrics in Table~\ref{tab:cls_ner_std}. 
The main observation is that AIA$^{2}$ maintains stable average performance while improving subgroup robustness. 
For Macro-F1 and Micro-F1, its standard deviation stays below 1.0 point on most datasets, indicating limited variation across seeds. 
Worst-group metrics show larger variance, as they depend on the most difficult subgroup and are more sensitive to small sample changes. 
Even so, AIA$^{2}$ remains more stable than JTT on all worst-group metrics; for example, its standard deviation is 1.24 on WWW2015 and 1.94 on RE3D, compared with 2.44 and 2.89 for JTT. 
These results indicate that iterative slice discovery and LLM-based augmentation do not introduce excessive seed sensitivity.

\begin{table}[t]
\centering
\footnotesize
\renewcommand{\arraystretch}{1.08}
\setlength{\tabcolsep}{3.5pt}

\resizebox{\columnwidth}{!}{%
\begin{tabular}{l ccc ccc}
\toprule
\multirow{2}{*}{\textbf{Setting}} &
\multicolumn{3}{c}{\textbf{HumAID}} &
\multicolumn{3}{c}{\textbf{CrossNER}} \\
\cmidrule(lr){2-4} \cmidrule(lr){5-7}
& \multicolumn{1}{c}{M-F1}
& \multicolumn{1}{c}{$\mu$-F1}
& \multicolumn{1}{c}{WGA}
& \multicolumn{1}{c}{M-F1}
& \multicolumn{1}{c}{$\mu$-F1}
& \multicolumn{1}{c}{WG-F1} \\
\midrule
Semantic-only   & 68.0 & 76.5 & 70.0 & 72.2 & 72.3 & 62.5 \\
$\alpha=0.05$   & 68.8 & 77.2 & 71.0 & 73.2 & 73.5 & 63.9 \\
$\alpha=0.1$    & 69.4 & \textbf{77.6} & 71.8 & \textbf{74.4} & \textbf{74.9} & 63.8 \\
$\alpha=0.2$    & 69.1 & 77.0 & \textbf{72.1} & \textbf{74.4} & 74.1 & \textbf{64.0} \\
$\alpha=0.5$    & \textbf{69.7} & 76.5 & 71.8 & 73.2 & \textbf{74.9} & 62.4 \\
$\alpha=1.0$    & 66.5 & 75.2 & 68.5 & 72.8 & 73.2 & 62.9 \\
Prediction-only & 64.2 & 73.1 & 65.5 & 70.5 & 71.8 & 61.8 \\
\bottomrule
\end{tabular}
}
\caption{
Sensitivity analysis of the weighting coefficient $\alpha$ in the joint semantic--predictive representation.
}
\label{tab:sensitivity_alpha}
\end{table}

\begin{table}[t]
\centering
\footnotesize
\renewcommand{\arraystretch}{1.08}
\setlength{\tabcolsep}{3.5pt}

\resizebox{\columnwidth}{!}{%
\begin{tabular}{c ccc ccc}
\toprule
\multirow{2}{*}{\textbf{$k$ in kNN graph}} &
\multicolumn{3}{c}{\textbf{HumAID}} &
\multicolumn{3}{c}{\textbf{CrossNER}} \\
\cmidrule(lr){2-4} \cmidrule(lr){5-7}
& \multicolumn{1}{c}{M-F1}
& \multicolumn{1}{c}{$\mu$-F1}
& \multicolumn{1}{c}{WGA}
& \multicolumn{1}{c}{M-F1}
& \multicolumn{1}{c}{$\mu$-F1}
& \multicolumn{1}{c}{WG-F1} \\
\midrule
5  & 68.6 & 77.4 & 70.1 & 72.8 & 74.0 & 62.2 \\
10 & 68.9 & 77.2 & 71.4 & 73.6 & 73.9 & 63.3 \\
15 & 69.4 & \textbf{77.6} & \textbf{71.8} & \textbf{74.4} & \textbf{74.9} & \textbf{63.8} \\
20 & \textbf{69.7} & 77.5 & 71.6 & \textbf{74.4} & 74.5 & 63.2 \\
\bottomrule
\end{tabular}
}
\caption{
Sensitivity analysis of the number of nearest neighbors in the kNN graph.
}
\label{tab:sensitivity_knn}
\end{table}

\begin{table}[t]
\centering
\footnotesize
\renewcommand{\arraystretch}{1.08}
\setlength{\tabcolsep}{3.5pt}

\resizebox{\columnwidth}{!}{%
\begin{tabular}{c ccc ccc}
\toprule
\multirow{2}{*}{\textbf{Resolution}} &
\multicolumn{3}{c}{\textbf{HumAID}} &
\multicolumn{3}{c}{\textbf{CrossNER}} \\
\cmidrule(lr){2-4} \cmidrule(lr){5-7}
& \multicolumn{1}{c}{M-F1}
& \multicolumn{1}{c}{$\mu$-F1}
& \multicolumn{1}{c}{WGA}
& \multicolumn{1}{c}{M-F1}
& \multicolumn{1}{c}{$\mu$-F1}
& \multicolumn{1}{c}{WG-F1} \\
\midrule
0.50 & 68.4 & 77.3 & 70.9 & 73.0 & 73.7 & 62.9 \\
0.75 & \textbf{69.4} & 77.1 & 71.1 & 73.6 & \textbf{75.0} & 63.1 \\
1.00 & \textbf{69.4} & \textbf{77.6} & \textbf{71.8} & \textbf{74.4} & 74.9 & \textbf{63.8} \\
1.25 & 69.1 & 77.5 & 71.4 & 74.3 & 74.5 & 63.6 \\
1.50 & 69.0 & 77.4 & 70.6 & 73.2 & 74.2 & 62.4 \\
\bottomrule
\end{tabular}
}
\caption{
Sensitivity analysis of the Leiden resolution parameter.
}
\label{tab:sensitivity_resolution}
\end{table}

\begin{table}[t]
\centering
\footnotesize
\renewcommand{\arraystretch}{1.08}
\setlength{\tabcolsep}{3.5pt}

\resizebox{\columnwidth}{!}{%
\begin{tabular}{c ccc ccc}
\toprule
\multirow{2}{*}{\textbf{\#Candidates}} &
\multicolumn{3}{c}{\textbf{HumAID}} &
\multicolumn{3}{c}{\textbf{CrossNER}} \\
\cmidrule(lr){2-4} \cmidrule(lr){5-7}
& \multicolumn{1}{c}{M-F1}
& \multicolumn{1}{c}{$\mu$-F1}
& \multicolumn{1}{c}{WGA}
& \multicolumn{1}{c}{M-F1}
& \multicolumn{1}{c}{$\mu$-F1}
& \multicolumn{1}{c}{WG-F1} \\
\midrule
10 & 68.4 & 77.5 & 70.8 & 72.9 & 74.2 & 62.5 \\
15 & 69.2 & 77.3 & 71.3 & 73.8 & 73.6 & 63.4 \\
20 & 69.4 & \textbf{77.6} & \textbf{71.8} & 74.4 & 74.9 & \textbf{63.8} \\
30 & \textbf{69.9} & \textbf{77.6} & 71.5 & \textbf{74.5} & \textbf{75.0} & 63.5 \\
\bottomrule
\end{tabular}
}
\caption{
Sensitivity analysis of the number of generated candidates per seed sample.
}
\label{tab:sensitivity_candidates}
\end{table}

\begin{table}[ht]
\centering
  
\resizebox{\columnwidth}{!}{%
\begin{tabular}{l cc cc}
\toprule
\multirow{2}{*}{\textbf{Baselines}} 
& \multicolumn{2}{c}{\textbf{WWW2015}} 
& \multicolumn{2}{c}{\textbf{CrossNER}} \\
\cmidrule(lr){2-3} \cmidrule(lr){4-5}
& Time (s) & Memory (GB) & Time (s) & Memory (GB) \\
\midrule
Focal    & 1143 & 17.6 & 1292   & 16.2 \\
GroupDRO & 1262 & 17.2 & 1666 & 18.6 \\
JTT      & 2837 & 21.0 & 2389 & 23.8 \\
DPE      & 1755 & 20.6 & 1973   & 21.0 \\
CB-LLM   & 3193 & 74.2 & 3372 & 78.0 \\
AugGPT   & 3760 & 72.4 & 3789   & 75.4 \\
AIA$^{2}$ & 2699 & 71.8 & 3022 & 76.6 \\
\bottomrule
\end{tabular}
}
\caption{Efficiency comparison on different datasets.}
\label{tab:efficiency_comparison}
\end{table}

\section{Hyperparameter Sensitivity Analysis}
\label{sec:hyperparameter_sensitivity}
{
We conduct a sensitivity analysis of AIA$^{2}$ for four key hyperparameters: the weighting coefficient $\alpha$ in the joint semantic--predictive representation, the number of nearest neighbors $k$ in the kNN graph, the Leiden resolution parameter, and the number of generated candidates per seed sample. 
The analysis is performed on two representative datasets: HumAID for CLS and CrossNER for NER. 
These hyperparameters cover the two main stages of AIA$^{2}$: latent slice discovery and LLM-based augmentation. 
For each experiment, we vary one hyperparameter while keeping all other settings fixed. 
The default setting used in the main experiments is $\alpha=0.1$, $k=15$, resolution $=1.0$, and $\#Candidates=20$.
}

{
Tables~\ref{tab:sensitivity_alpha}--\ref{tab:sensitivity_candidates} show that AIA$^{2}$ is stable under moderate hyperparameter changes. 
For $\alpha$, semantic-only corresponds to $\alpha=0$, whereas prediction-only represents the $\alpha\rightarrow\infty$ limit. 
Values between $0.05$ and $0.2$ outperform semantic-only slicing across all metrics on both datasets. 
The default $\alpha=0.1$ remains competitive across tasks, while larger values yield less consistent results and prediction-only performs worst overall. 
For $k$, both datasets remain stable across $k \in \{5,10,15,20\}$; the default value $k=15$ gives the best or near-best worst-group performance, while $k=20$ slightly improves Macro-F1. 
For the Leiden resolution parameter, resolution $=1.0$ gives the best results on HumAID and the best worst-group score on CrossNER. 
The tested resolutions lead to small performance changes, showing that AIA$^{2}$ does not rely on a narrow clustering granularity. 
For the number of generated candidates, a larger candidate pool improves Macro-F1 in some cases, but $\#Candidates=20$ gives the best worst-group performance on both datasets. 
Overall, the default setting remains competitive across tasks and usually provides the strongest worst-group results. 
These results suggest that the gains of AIA$^{2}$ are not driven by dataset-specific tuning or a fragile hyperparameter choice.
}

\section{Computational Overhead}
\label{sec:computational}
{
We report training time and peak GPU-memory cost on one representative dataset per task: WWW2015 for CLS and CrossNER for NER. 
All methods are evaluated with batch size 16 on an H100 GPU. 
Table~\ref{tab:efficiency_comparison} shows that AIA$^{2}$ introduces additional overhead over lightweight training baselines such as Focal Loss and GroupDRO, due to latent slice discovery and LLM-based targeted augmentation. 
However, AIA$^{2}$ is more efficient than generic LLM augmentation baselines in training time. 
On WWW2015, it reduces training time by 15.5\% compared with CB-LLM and 28.2\% compared with AugGPT; on CrossNER, it reduces training time by 10.4\% and 20.2\%, respectively. 
Its peak memory cost is also comparable to LLM augmentation baselines: AIA$^{2}$ uses less memory than CB-LLM on both datasets, less than AugGPT on WWW2015, and slightly more than AugGPT on CrossNER. 
These results show that AIA$^{2}$ provides a practical cost--benefit tradeoff: it adds expected overhead over simple training baselines while remaining more time-efficient than general-purpose LLM augmentation methods.
}

\begin{figure*}[!t]
\centering
\begin{minipage}{0.462\textwidth}
    \centering
    \includegraphics[width=\linewidth]{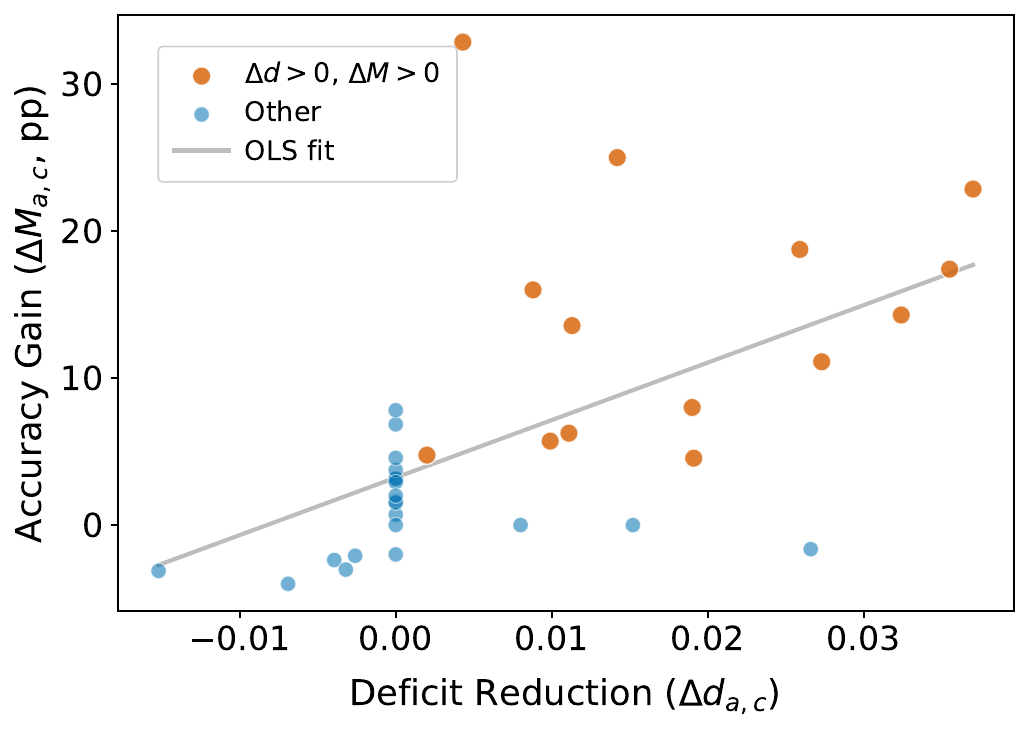}
    \vspace{-23pt}
    \caption*{(a) HumAID}
\end{minipage}
\hfill
\begin{minipage}{0.462\textwidth}
    \centering
    \includegraphics[width=\linewidth]{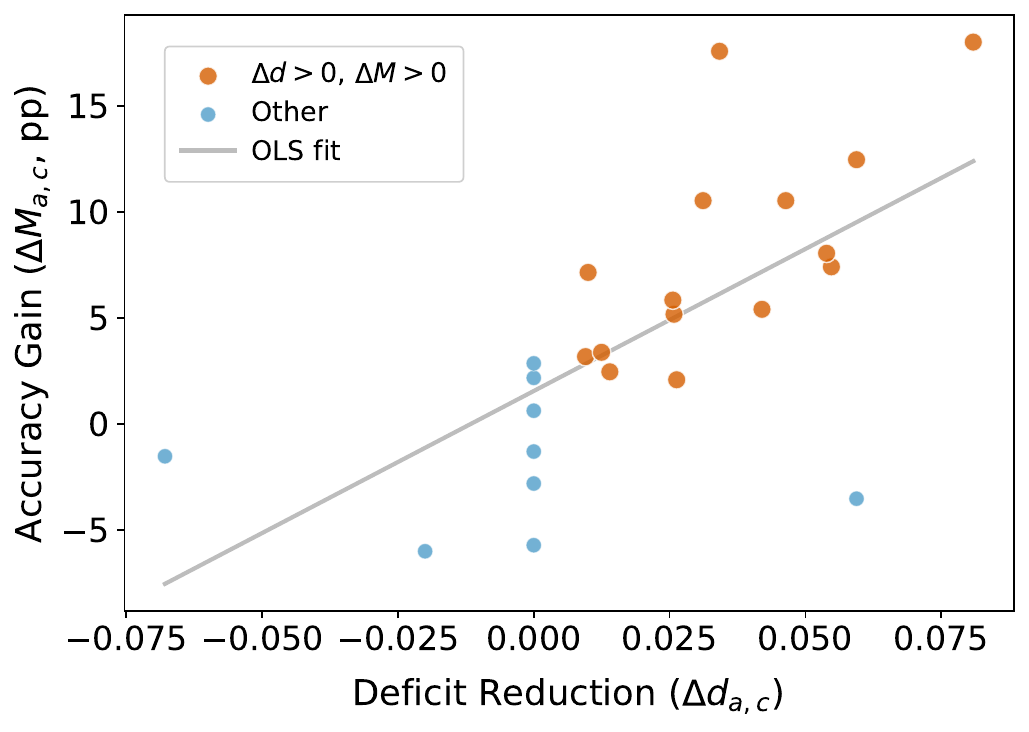}
    \vspace{-23pt}
    \caption*{(b) HateXplain}
\end{minipage}
\vspace{-6pt}
\caption{
Additional subgroup-label accuracy gain versus deficit reduction on HumAID and HateXplain. 
}
\label{fig:cls_deficit_humaid_hatexplain}
\end{figure*}

\section{Additional Generated Data Analysis}
\label{sec:additional_generated_analysis}

To complement the WWW2015 analysis in Section~\ref{sec:slice_deficit_analysis}, we further examine the relationship between deficit reduction and subgroup-label accuracy gain on the other two CLS datasets, HumAID and HateXplain. 
As shown in Figure~\ref{fig:cls_deficit_humaid_hatexplain}, both datasets show a clear positive association between the reduction of subgroup-label deficits and the improvement of subset-level accuracy. 
On HumAID, the trend is significantly positive across 35 subgroup-label subsets, with Spearman's $\rho=0.700$ ($p=2.93\times10^{-6}$) and an OLS fit of $R^2=0.322$. 
On HateXplain, the association is also significant across 24 subsets, with Spearman's $\rho=0.711$ ($p=9.72\times10^{-5}$) and $R^2=0.427$. 
In addition, 14 subsets on HumAID and 15 subsets on HateXplain fall into the positive-repair region, where deficit reduction and accuracy gain are both positive. 
These results are consistent with the main WWW2015 finding and further support that AIA$^{2}$ improves subgroup robustness by repairing local subgroup-label deficits, rather than relying only on untargeted data expansion.

\begin{table}[t]
\centering
\footnotesize
\renewcommand{\arraystretch}{1.08}
\setlength{\tabcolsep}{3.5pt}

\resizebox{\columnwidth}{!}{%
\begin{tabular}{l cc cc}
\toprule
\multirow{2}{*}{\textbf{Method}} &
\multicolumn{2}{c}{\textbf{HumAID}} &
\multicolumn{2}{c}{\textbf{CrossNER}} \\
\cmidrule(lr){2-3} \cmidrule(lr){4-5}
& \multicolumn{1}{c}{Added}
& \multicolumn{1}{c}{Increase}
& \multicolumn{1}{c}{Added}
& \multicolumn{1}{c}{Increase} \\
\midrule
AugGPT              & 1,493 & 13.9\% & 2,196 & 51.5\% \\
CB-LLM              &   997 &  9.3\% & 1,913 & 44.9\% \\
Semantic Cluster-Gap & 1,281 & 12.0\% & 1,952 & 45.8\% \\
AIA$^{2}$           & 1,073 & 10.0\% & 1,643 & 38.6\% \\
\bottomrule
\end{tabular}
}
\caption{
Retained augmentation volume across generative methods on HumAID and CrossNER.
}
\label{tab:augmentation_volume}
\end{table}

Table~\ref{tab:augmentation_volume} reports the filtered samples retained for training; raw candidates rejected during quality control are excluded, and the increase is measured relative to the original training-set size.
Semantic Cluster-Gap corresponds to the semantic-only setting in Table~\ref{tab:sensitivity_alpha}: it forms semantic communities and targets labels that are underrepresented relative to the global label distribution.
AIA$^{2}$ retains 1,073 samples on HumAID and 1,643 on CrossNER, corresponding to training-set increases of 10.0\% and 38.6\%, respectively.
It retains fewer samples than AugGPT and Semantic Cluster-Gap on both datasets and fewer than CB-LLM on CrossNER, while CB-LLM retains slightly fewer samples on HumAID.
These comparisons indicate that the performance gains of AIA$^{2}$ do not arise from consistently using a larger augmentation volume.

\begin{table}[t]
\centering
\footnotesize
\renewcommand{\arraystretch}{1.08}
\setlength{\tabcolsep}{3.5pt}

\resizebox{\columnwidth}{!}{%
\begin{tabular}{l ccc ccc}
\toprule
\multirow{2}{*}{\textbf{Setting}}
& \multicolumn{3}{c}{\textbf{WWW2015}}
& \multicolumn{3}{c}{\textbf{CrossNER}} \\
\cmidrule(lr){2-4}\cmidrule(lr){5-7}
& M-F1 & $\mu$-F1 & WGA
& M-F1 & $\mu$-F1 & WG-F1 \\
\midrule
Full AIA$^{2}$
& \textbf{57.6} & \textbf{87.5} & \textbf{80.9}
& \textbf{74.4} & \textbf{74.9} & \textbf{63.8} \\
Semantic-only slicing
& 55.4 & 85.3 & 77.7
& 72.2 & 72.3 & 62.5 \\
Prediction-only slicing
& 54.3 & 85.8 & 76.6
& 70.5 & 71.8 & 61.8 \\
\addlinespace[2pt]
Deficit-only selection
& 56.1 & 86.4 & 79.0
& 72.5 & 73.7 & 62.8 \\
High-loss-only selection
& 56.5 & 86.3 & 78.2
& 72.0 & 73.3 & 62.6 \\
\bottomrule
\end{tabular}%
}

\caption{
Ablations of slice representation and seed selection.
}
\label{tab:additional_ablations}
\end{table}

\begin{table*}[!t]
\centering
\footnotesize
\renewcommand{\arraystretch}{1.15}
\setlength{\tabcolsep}{3.0pt}

\resizebox{0.86\textwidth}{!}{%
\begin{tabular}{l *{15}{c}}
\toprule
\multirow{3}{*}{\textbf{Method}}
& \multicolumn{9}{c}{\textbf{CLS}}
& \multicolumn{6}{c}{\textbf{NER}} \\
\cmidrule(lr){2-10}\cmidrule(lr){11-16}
& \multicolumn{3}{c}{\textbf{HumAID}}
& \multicolumn{3}{c}{\textbf{WWW2015}}
& \multicolumn{3}{c}{\textbf{HateXplain}}
& \multicolumn{3}{c}{\textbf{RE3D}}
& \multicolumn{3}{c}{\textbf{CrossNER}} \\
\cmidrule(lr){2-4}\cmidrule(lr){5-7}\cmidrule(lr){8-10}\cmidrule(lr){11-13}\cmidrule(lr){14-16}
& M-F1 & $\mu$-F1 & WGA
& M-F1 & $\mu$-F1 & WGA
& M-F1 & $\mu$-F1 & WGA
& M-F1 & $\mu$-F1 & WG-F1
& M-F1 & $\mu$-F1 & WG-F1 \\
\midrule
Leiden        & \textbf{69.4} & \textbf{77.6} & \textbf{71.8} & \textbf{57.6} & \textbf{87.5} & \textbf{80.9} & \textbf{64.6} & \textbf{66.2} & \textbf{57.7} & \textbf{66.5} & \textbf{66.3} & \textbf{60.5} & \textbf{74.4} & \textbf{74.9} & 63.8 \\
K-Means       & \textbf{69.4} & \textbf{77.6} & 71.7 & 55.4 & 87.1 & 80.3 & 63.4 & 65.0 & \textbf{57.7} & 63.9 & 63.4 & 55.3 & 73.0 & 72.5 & 61.8 \\
DBSCAN        & 68.1 & 77.2 & 71.1 & 55.7 & 86.6 & \textbf{80.9} & 62.6 & 64.2 & 55.8 & 60.8 & 61.5 & 55.7 & 72.2 & 70.6 & 60.4 \\
Agglomerative & 68.4 & 76.9 & 70.7 & 52.7 & 86.1 & 79.4 & 61.5 & 63.4 & 55.8 & 62.9 & 63.5 & 59.3 & \textbf{74.3} & 73.1 & \textbf{64.1} \\
\bottomrule
\end{tabular}%
}

\caption{
Clustering method robustness for latent slice discovery.
Leiden is the default clustering method, while K-Means, DBSCAN, and Agglomerative clustering are used as alternatives.
}
\label{tab:clustering_robustness}
\end{table*}

\section{Additional Ablations}
\label{app:additional_ablations}

\paragraph{Representation and selection}
We further examine the representations used for latent slice discovery and the signals used for seed selection on WWW2015 and CrossNER, while keeping the remaining pipeline and augmentation budget unchanged.
Semantic-only and prediction-only slicing use only semantic embeddings or model-output distributions, respectively.
Deficit-only selection randomly samples within deficit labels, whereas high-loss-only selection ranks all instances within each slice without using deficit labels.
As shown in Table~\ref{tab:additional_ablations}, the full setting achieves the best results across all metrics on both datasets.
It improves worst-group performance over the single-signal slice variants by 3.2--4.3 points on WWW2015 and 1.3--2.0 points on CrossNER, and over the single-signal selection variants by 1.9--2.7 and 1.0--1.2 points, respectively.
These results support the joint designs used in both stages.

\paragraph{Clustering methods}
We include this experiment to examine whether the effectiveness of AIA$^{2}$ depends on the specific clustering algorithm used for latent slice discovery.
Table~\ref{tab:clustering_robustness} compares the default Leiden community detection method~\cite{traag2019leiden} with K-Means~\cite{Kmeans2023Abiodun}, DBSCAN~\cite{Ester1996density}, and Agglomerative clustering~\cite{Mllner2011ModernHA}.
Overall, Leiden provides the most consistent performance across datasets.
K-Means performs comparably on classification tasks but drops more clearly on NER, suggesting that centroid-based clustering may be less effective for irregular error regions.
DBSCAN is less stable, likely because the joint semantic--predictive space has varying local densities.
Agglomerative clustering shows mixed results, with occasional gains on CrossNER but weaker performance on several other datasets.
These results suggest that AIA$^{2}$ is broadly robust to reasonable clustering choices, while graph-based community detection remains the most reliable default for iterative slice-aware augmentation.

\end{document}